\documentclass[lettersize,journal,twoside]{IEEEtran}
\usepackage{cite}

\usepackage{multirow}
\usepackage{hyperref}
\usepackage{stackengine}
\usepackage{amsmath}
\usepackage{amsfonts}
\usepackage{amssymb}
\usepackage{booktabs,caption}
\usepackage[flushleft]{threeparttable}
\usepackage{subcaption}
\usepackage[ruled,vlined]{algorithm2e}

\usepackage{arydshln}
\usepackage{float}
\floatstyle{plaintop}
\restylefloat{table}

\usepackage[pdftex]{graphicx}
\graphicspath{ {./figures/} }
\DeclareGraphicsExtensions{.pdf,.jpg,.png}

\usepackage{multirow}

\usepackage{fancyhdr}
\usepackage{kantlipsum}
\usepackage{eso-pic}

\begin{document}

\AddToShipoutPictureBG*{%
	\AtPageLowerLeft{%
		\setlength\unitlength{1in}%
		\hspace*{\dimexpr0.5\paperwidth\relax}
		\makebox(0,0.63)[c]{2379-8858~\copyright2022 IEEE. This work has been accepted for publication in IEEE Transactions on Intelligent Vehicles.}
		\makebox(0,0.3)[c]{The published version can be accessed at \href{https://ieeexplore.ieee.org/document/9802907}{https://ieeexplore.ieee.org/document/9802907}. DOI: \href{https://doi.org/10.1109/TIV.2022.3185303}{10.1109/TIV.2022.3185303}}
}}

\title{End-to-end Autonomous Driving with\\Semantic Depth Cloud Mapping and Multi-agent}

\author{
Oskar~Natan
and~Jun~Miura,~\IEEEmembership{Member,~IEEE}
\thanks{Manuscript received April 8, 2022; revised May 10, 2022 and May 27, 2022; Accepted June 19, 2022. The Associate Editor for this article is .... Corresponding author: Oskar Natan.} %
\thanks{Oskar Natan is with the Department of Computer Science and Engineering, Toyohashi University of Technology, Toyohashi, Aichi 441-8580, Japan, and also with the Department of Computer Science and Electronics, Gadjah Mada University, Yogyakarta 55281, Indonesia (e-mail: oskar.natan.ao@tut.jp; oskarnatan@ugm.ac.id).}
\thanks{Jun Miura is with the Department of Computer Science and Engineering, Toyohashi University of Technology, Toyohashi, Aichi 441-8580, Japan (e-mail: jun.miura@tut.jp).}
}

\markboth{IEEE Transactions on Intelligent Vehicles,~Vol.~xx, No.~x, Mm~yyyy}{Natan and Miura: End-to-end Autonomous Driving with Semantic Depth Cloud Mapping and Multi-agent}


\maketitle


\begin{abstract}
	Focusing on the task of point-to-point navigation for an autonomous driving vehicle, we propose a novel deep learning model trained with end-to-end and multi-task learning manners to perform both perception and control tasks simultaneously. The model is used to drive the ego vehicle safely by following a sequence of routes defined by the global planner. The perception part of the model is used to encode high-dimensional observation data provided by an RGBD camera while performing semantic segmentation, semantic depth cloud (SDC) mapping, and traffic light state and stop sign prediction. Then, the control part decodes the encoded features along with additional information provided by GPS and speedometer to predict waypoints that come with a latent feature space. Furthermore, two agents are employed to process these outputs and make a control policy that determines the level of steering, throttle, and brake as the final action. The model is evaluated on CARLA simulator with various scenarios made of normal-adversarial situations and different weathers to mimic real-world conditions. In addition, we do a comparative study with some recent models to justify the performance in multiple aspects of driving. Moreover, we also conduct an ablation study on SDC mapping and multi-agent to understand their roles and behavior. As a result, our model achieves the highest driving score even with fewer parameters and computation load. To support future studies, we share our codes at \href{https://github.com/oskarnatan/end-to-end-driving}{https://github.com/oskarnatan/end-to-end-driving}. 
\end{abstract}


\section{Introduction}

\IEEEPARstart{A}{utonomous} driving is a complex intelligent system consisting of several subsystems that handle multiple subtasks from perception phase to control phase\cite{intro0}\cite{AD_review1}. The solution for each task can be done by simply employing a specific module\cite{intro1}. However, this approach is costly and inefficient as a further configuration is needed to form an integrated modular system\cite{intro2}. For example, we must assign the information provided by perception modules (e.g., semantic segmentation, object detection) as the input for the controller module. This integration process can be very tedious and may lead to information loss as a lot of parameters adjustment is done manually. With rapid deep learning research, plenty of works have been conducted to address this issue by training a single model in end-to-end and multi-task manners\cite{intro3}\cite{intro4}\cite{intro5}. With this strategy, the model can be trained to provide the final action solely based on the observation data captured with a set of sensors. As manual tuning is no longer needed, the model can leverage the extracted features all by itself. Inside the model, the controller module can take the information given by the perception module continuously\cite{intro6}. 

\begin{figure}[!t]
	\centering
	\includegraphics[width=\linewidth]{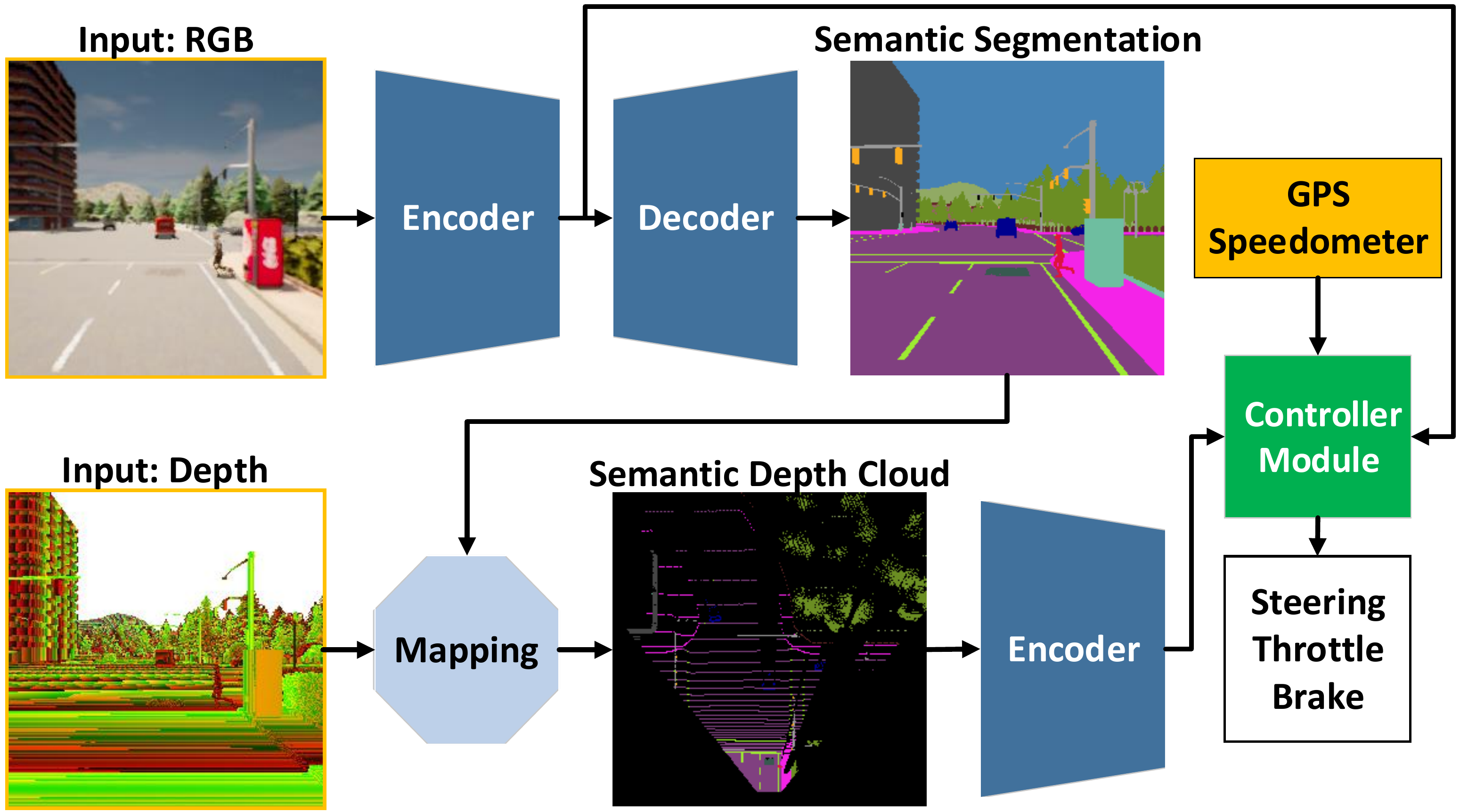}
	\caption{Proposed end-to-end multi-task model. The model takes a set of inputs provided by RGBD camera, GPS, and speedometer to drive the ego vehicle. The details of the network architecture can be seen on Figure \ref{fig:netarch}.} %
	\label{fig:overview}
\end{figure}

Currently, the challenge that remains for an end-to-end model is how to encode or extract good features so that the controller module can decode them into proper vehicular controls. Obviously, the perception module needs to be supported with many kinds of information that represent the detail condition around the ego vehicle. In this case, sensor fusion-based models have been proven to achieve better performance as it uses various kinds of sensors\cite{intro7}\cite{intro8}\cite{intro7a}\cite{intro8a}. However, a huge computation load is inevitable as a bigger model is needed to process large data. Moreover, data preprocessing technique is necessary as varying sensors often come with different data modalities\cite{intro9}\cite{intro10}. Even with a good set of encoded features, there is still another challenge that remains where the controller module can be stuck with a certain behavior due to its learning experience. Therefore, more decision-makers may be needed to translate extracted information into a different aspect of driving control\cite{intro11}\cite{intro12}. Furthermore, an imbalance learning during the training process could be another issue since both perception and control tasks are done simultaneously. Hence, a method to balance the training signal is also necessary to ensure that all tasks are learned at the same pace\cite{intro13}\cite{intro14}\cite{intro15}.

To answer those challenges, we propose an end-to-end deep multi-task learning model as shown in Figure \ref{fig:overview}. We consider imitation learning, especially the behavior cloning approach as it allows the leveraging of large-scale datasets to be used to train the model to near-human standard\cite{imitation1}\cite{imitation2}. The model is made of two main modules, the perception module (blue) and the controller module (green). Our model takes RGB image and depth map of the front view as the input for its perception module. Thanks to the rapid development of sensor devices, both observation data can be provided with a single RGBD camera so that there is no need to mount more sensors on the ego vehicle\cite{intro16}\cite{intro17}\cite{intro18}\cite{intro19}. Besides that, RGB image and depth map also have the same data shape and representation so that both information can be processed easily. Meanwhile, the controller module is responsible for decoding the extracted features from the perception module along with additional information of route location and measured speed provided by GPS and speedometer. By using two different agents, more varied final control actions can be made considering multiple aspects of driving\cite{intro20}\cite{intro21}\cite{intro22}. Furthermore, we design the model to only have a small number of neurons or trainable parameters to reduce the computation load. Finally, we use an adaptive loss weighting algorithm called modified gradient normalization (MGN) to balance the training signal\cite{mgn}. Thus, the imbalance learning problem can be solved and the model does not tend to focus only on a single task during the training process. The novelty of this research is summarized as follows:
\begin{itemize}
	\item We present an end-to-end deep multi-task learning model that takes a set of inputs provided by an RGBD camera, GPS, and speedometer to perform both perception and control tasks in one forward pass. To be more specific, the perception module leverages the use of the bird's eye view (BEV) semantic depth cloud (SDC) mapping to enhance its capability for scene understanding, especially in distinguishing free and occupied areas. In the controller module, the model has two different agents that determine the final actions for driving the vehicle. A more varied decision allows the model to consider multiple aspects of good drivability.
	\item We conduct a comparative study with some recent models to justify model performance on the driving task and other task-specific evaluations. Based on the experimental results in four different scenarios, our model achieves the best performance even with fewer parameters. We also conduct an ablation study to understand the contribution of SDC mapping and multi-agent. 
\end{itemize}

The rest of this paper is organized as follows. In Section \ref{sec:related}, we do a comprehensive review of some related works that also inspire this research. In Section \ref{sec:model}, we explain our proposed model, especially about perception and control modules. Then, we explain the experiment, including the task, data generation, training configuration, and evaluation metrics in Section \ref{sec:exp_setup}. In Section \ref{sec:result}, we analyze the result to understand the model drivability and behavior in comparison with some recent models. Finally, we conclude our findings in Section \ref{sec:conclusion} along with some suggestion for future studies.

\section{Related Work} \label{sec:related}

In this section, we review some related works in the area of end-to-end autonomous driving with the imitation learning approach. Then, we explain how they inspire our research. 

\subsection{End-to-end Multi-task Model} \label{subs:related1}
There are two main advantages in training a model in end-to-end and multi-task learning manners. With an end-to-end learning fashion, there are no additional settings needed to integrate all submodules so that such kinds of information loss and human error can be avoided. Then, with a multi-task learning (MTL) strategy, the model can leverage shared features to speed up the training process. In the autonomous driving research, Ishihara et al.\cite{highcommand1} have demonstrated the usefulness of training a model in end-to-end and multi-task manners. Similar to CILRS\cite{cilrs} (conditional imitation learning with ResNet\cite{resnet} and speed input) model, their model takes front RGB image, speed measurement, and discrete high-level navigational command to predict the level of steering, throttle, and brake used to drive the vehicle. In addition, the MTL approach on depth estimation, semantic segmentation, and traffic light state prediction are used to improve the quality of extracted features in the perception module. With better features, the controller module is expected to be better at determining vehicular controls. A similar imitation learning-based approach has been studied by Chen et al.\cite{lbc} where the same set of inputs is used to drive the vehicle. However, instead of using discrete high-level navigational commands directly, the model produces a set of waypoints used by two PID controllers to drive the vehicle. 

In our research, we take the idea of performing multiple perception tasks of semantic segmentation and traffic light state prediction as extra supervision demonstrated by Ishihara et al.\cite{highcommand1} to guide the perception module in producing better features for the controller module. However, instead of performing depth estimation, we use a depth map provided by the RGBD camera as an input to the model which opens the possibility of sensor fusion strategy in performing better scene understanding\cite{intro7}\cite{intro10}\cite{unetblock2}.  Another issue that needs to be addressed in the multi-task learning approach is the imbalanced learning problem where the model may tend to focus on a certain task only\cite{intro13}\cite{intro14}\cite{intro15}. To address this issue, we use an adaptive learning algorithm called modified gradient normalization to ensure that all tasks are learned at the same pace\cite{mgn}.

\begin{figure*}
	\centering
	\includegraphics[width=\textwidth]{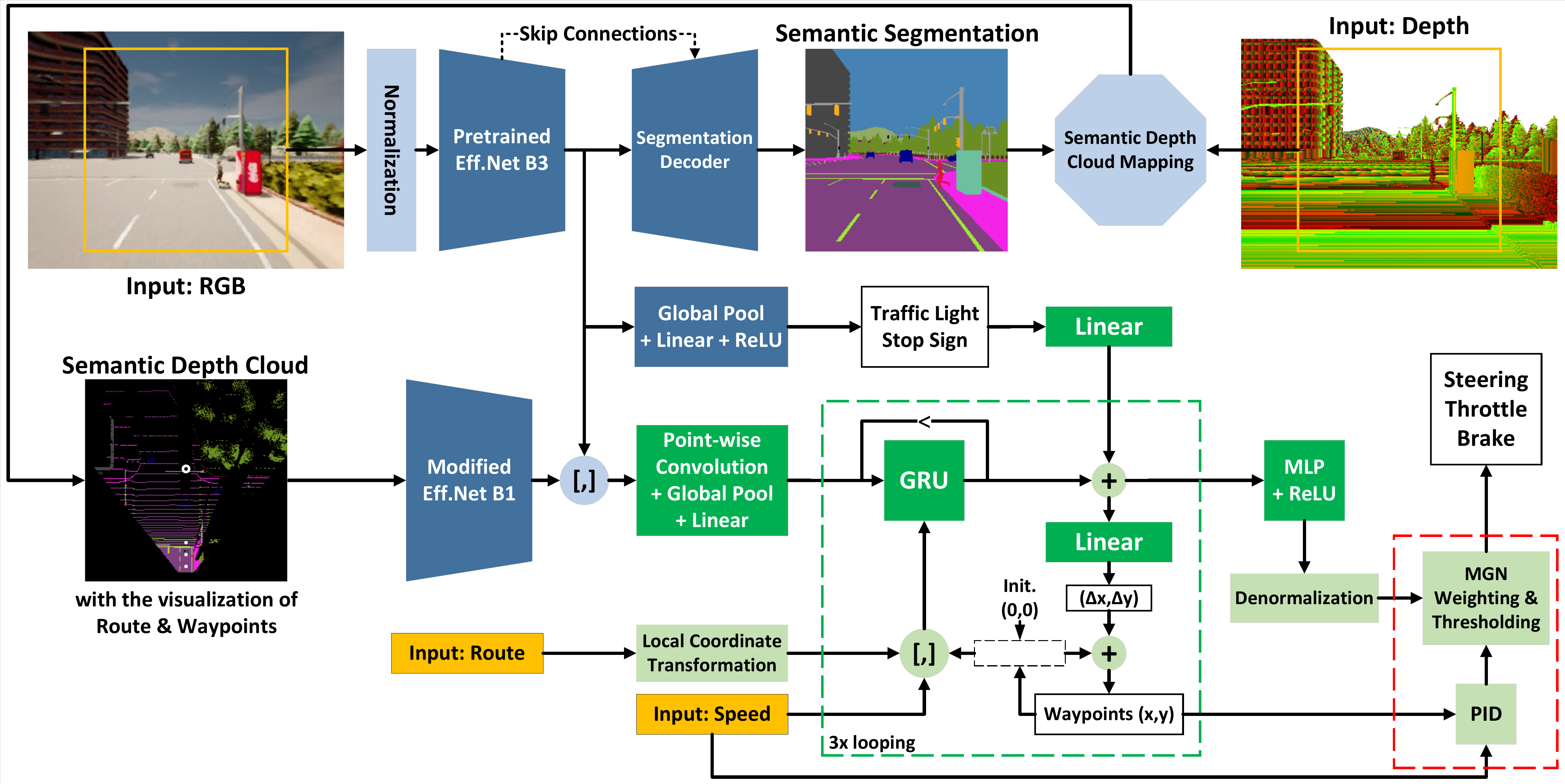}
	\caption{Proposed model. The light-colored items are not trainable. Blue colored items are considered as the perception module while the green colored items are the controller module. The process inside the dashed green line box is looped over three times. Meanwhile, items inside the dashed red line box are only used for driving. Therefore, the model predicts waypoints and estimates the level of steering, throttle, and brake separately during the training process. Inside semantic depth cloud, the route is represented with a white hollow circle, while the waypoints are represented with small white circles.}
	\label{fig:netarch}
\end{figure*}

\subsection{Multi-modal Sensor Fusion Strategy}\label{subs:related2}
By using the sensor fusion technique, a model can have a better scene understanding as it opens plenty of possibilities to perceive the environment in multiple kinds of representation. In the field of autonomous driving research, Huang et al.\cite{highcommand0} have proposed a sensor fusion-based model that takes an RGB image and depth map to capture a deeper global context in front of the vehicle. Both inputs are fused at an early stage to form low-dimensional latent features. The extracted features are processed by the controller module to determine a set of actions. Similar to Ishihara et al.\cite{highcommand1} and Chen et al.\cite{lbc}, this approach also uses navigational commands to drive the vehicle. The sensor fusion technique also opens the possibilities of perceiving the environment from a different perspective. Prakash et al.\cite{transfuser} has developed a sensor fusion-based model that takes an RGB image and preprocessed LiDAR point clouds. The RGB image contains the information of the front-view perspective while the LiDAR contains the information of the top view or bird's eye view (BEV) perspective. By fusing both features, the model can perform a better scene understanding\cite{lidarpreprocessed0}\cite{lidarpreprocessed1}. Moreover, unlike most current works, this model does not use high-level navigational commands to drive, instead, it takes sparse GPS location of predefined routes provided by a global planner.

In our research, we also use a combination of an RGB image and depth map provided by a single RGBD camera. However, instead of extracting depth features at the early stage, we project the depth map and perform semantic depth cloud (SDC) mapping with BEV perspective. Therefore, the model can take the advantage of perceiving the environment from the top view perspective\cite{transfuser}\cite{lidarpreprocessed0}\cite{lidarpreprocessed1}. Moreover, since the SDC stores semantic information, the model can learn a better understanding as the occupied or drivable regions become clearer than preprocessed LiDAR point clouds which only contain height information. We also consider using a sequence of routes instead of high-level navigational commands as it makes more sense for driving an autonomous vehicle in real-world condition\cite{use_route1}\cite{use_route2}\cite{use_route3}.

\section{Proposed Model}\label{sec:model}
In this section, we describe the details of the network architecture. As shown in Figure \ref{fig:netarch}, the model is composed of two main modules, the perception module (blue) and the controller module (green). Concisely, the perception module is responsible for complex scene understanding and providing useful information to the control module. To be more specific, the perception module performs semantic segmentation, semantic depth cloud (SDC) mapping in a bird's eye view (BEV) perspective, traffic light state prediction, and stop sign prediction. Then, the controller module leverage the given information in the form of encoded features together with additional inputs of current speed measurement and the GPS location of the route. This module provides waypoints and vehicular controls as the final outputs.

\subsection{Perception Module}\label{subs:perception}

\begin{figure*}
	\centering
	\includegraphics[width=\textwidth]{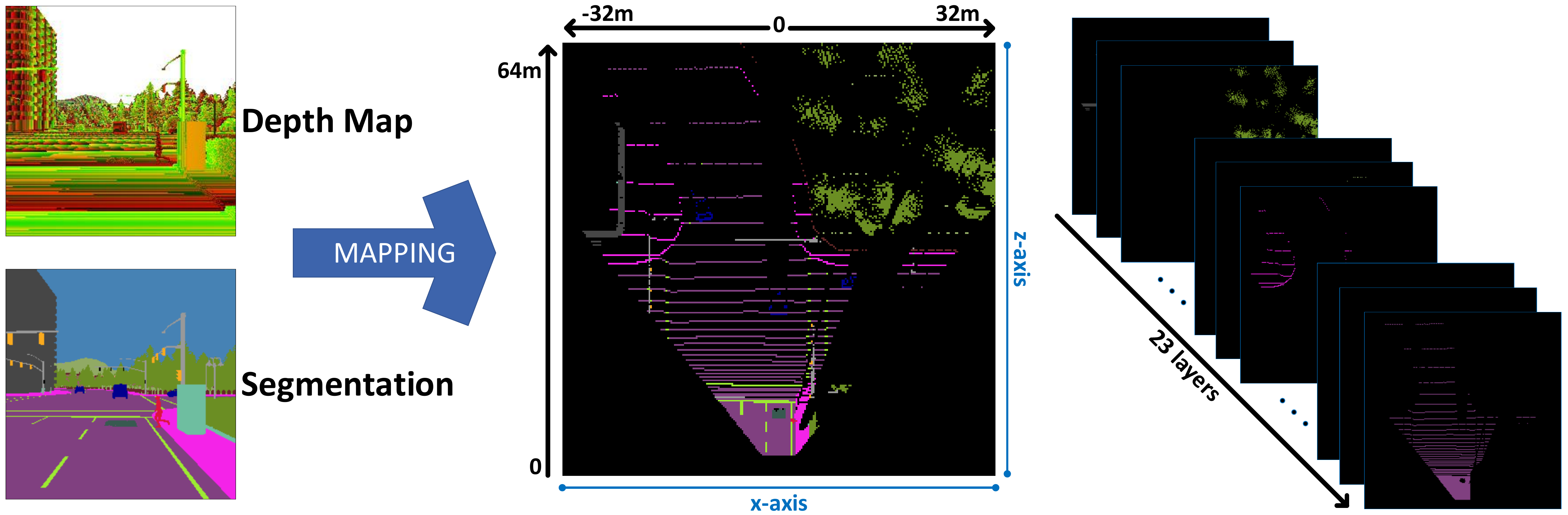}
	\caption{Semantic depth cloud mapping. Each layer holds one unique class mentioned in Table \ref{tab:data_info}.}
	\label{fig:sdc_parts}
\end{figure*}

The perception module takes an RGB image and depth map provided by a single RGBD camera as its main inputs. We consider a region of interest (ROI) of $256\times256$ at the center for a fair comparison with other models (see Section \ref{sec:result} for more details). Another purpose is to eliminate the distortion at the corner that causes RGB image and depth map to have a different appearance. As shown in Figure \ref{fig:netarch}, we begin the perception phase with ImageNet normalization on the RGB image since we use EfficientNet version B3\cite{effnet} pre-trained on ImageNet\cite{imagenet} as the RGB encoder. EfficientNet is chosen as it can perform excellently on many vision-related tasks with a small number of parameters. Then, the extracted feature maps are learned by the segmentation decoder to perform semantic segmentation in 23 different classes mentioned in Table \ref{tab:data_info}. The decoder is made of multiple convolution blocks ($2\times$($3\times3$ convolution $+$ batch normalization\cite{batchnorm} $+$ ReLU\cite{relu}) $+$ bilinear interpolation) and a final pointwise $1\times1$ convolution with sigmoid activation. It is also enhanced with different scales of extracted feature maps from the encoder connected with some skip connections\cite{unetblock2}\cite{unetblock}. In addition, we also create a separate module consisting of a global average pooling layer, linear layer, and ReLU to specifically predict the traffic light state and stop sign. Although both tasks can be considered classification problems, we choose ReLU rather than sigmoid since we want to avoid information loss as the prediction outputs will be encoded later in the controller module for extra supervision.

In addition to the front view perspective, the scene understanding capability can be improved further by providing more information from the BEV perspective. Thus, the agent can have a better capability in estimating the drivable regions. Hence, we perform semantic depth cloud (SDC) mapping using segmentation prediction and depth map. However, we ignore the height information (y-axis) as we consider BEV perspective which relies on the x-axis and z-axis (given by depth map). Therefore, if there are multiple object classes corresponding to one point on the 2D plane, then the object that has the highest location is selected. The SDC mapping process can be summarized as follows:
\begin{itemize}
	\item Define the distance range of 64 meters to the front and 32 meters to the left and right forming a coverage area of $64 \times 64$ square meters. This means that the ego vehicle is always positioned at the bottom-center. Then, define the spatial dimension of the SDC tensor $H \times W = 256\times256$, so that one element is equal to an area of $25 \times 25$ square centimeters. 
	\item Get segmentation tensor $\mathcal{S}$, depth tensor $\mathcal{D}$, transformation matrix for x-axis $\mathcal{T}_x$ (formed with camera parameter $f_x$ and ROI size of $256\times256$).
	\item Compute x-axis ($\mathcal{P}_x$) and z-axis ($\mathcal{P}_z$) coordinates and normalize them to match the spatial dimension of the SDC tensor with (\ref{eq:px}) and (\ref{eq:zx}).
		\begin{equation} \label{eq:px}
			\mathcal{P}_x = \left\lfloor \frac{(\mathcal{D} \times \mathcal{T}_x + 32)}{64} \times 255 \right\rceil,
		\end{equation}
		\begin{equation} \label{eq:zx}
			\mathcal{P}_z = \left\lfloor \bigg(1 - \frac{\mathcal{D}}{64}\bigg) \times 255 \right\rceil,
		\end{equation}
		Keep in mind that tensor index start from 0, hence, we use 255 instead of 256. Unlike the standard cartesian coordinate system, any of (x,0) coordinates are located at the top of 2D plane. Thus, we do mirroring and shifting $(1 - \frac{\mathcal{D}}{64})$ on $\mathcal{P}_z$ computation. 
	\item Project every element containing a certain semantic class in $\mathcal{S}$ to a $256\times256$ matrix based on $\mathcal{P}_x$ and $\mathcal{P}_z$.
	\item Finally, apply one-hot encoding to the matrix to obtain SDC tensor $\mathbb{R} \in \{0,1\}^{23 \times 256 \times 256}$, where 23 is the channel representing the number of classes and $256 \times 256$ is the spatial dimension. Each channel holds one semantic class as shown in Figure \ref{fig:sdc_parts}.
	
\end{itemize}


 With this representation, the model can perceive better and drive the ego vehicle safely in the environment. We use a smaller version of EfficientNet named B1\cite{effnet} as the SDC encoder. There is no need to use the same or even a bigger model than the RGB encoder as the SDC already contains very distinctive information. Since the SDC contains 23 layers of semantic segmentation classes, we modify the first convolutional layer to receive a tensor with 23 channels. Then, the entire encoder is initialized with Kaiming initialization \cite{kaiming_init} to catch up with the RGB encoder in terms of convergence during the training process. Finally, both extracted RGB and SDC features are concatenated to form a tensor with the size of $C \times H \times W = (1536+1280) \times 8 \times 8$, where 1536 and 1280 are the number of channels in RGB and SDC features, respectively. To be noted, the RGB features have a greater number of channels than SDC features as it is extracted with a bigger EfficientNet model version. Meanwhile, the spatial dimension of $8 \times 8$ is a result of multiple downsampling through the EfficientNet architecture.

\subsection{Controller Module}\label{subs:controller}
The controller module is used to decode RGB and SDC features which have been concatenated in the perception module. We begin the control phase by employing a fusion block composed of pointwise $(1\times1)$ convolutional layer, global average pooling layer, and linear layer to process the features. The point-wise convolutional layer is used to fuse and learn the relation of each feature element along the channel axis and results in a smaller feature tensor with a size of $C \times H \times W = 384 \times 8 \times 8$. Then, the global average pooling layer is used to obtain the global context by averaging $8 \times 8$ array on each channel. Meanwhile, the linear layer is used to reduce 384 feature elements into 232 feature elements so that the upcoming layers do not need to process an enormous number of neurons to save computational load.

We use a gated recurrent unit (GRU) introduced by Cho et al.\cite{gru} to further process the features from the fusion block. In the network architecture, GRU is chosen as the model needs to keep relevant information for waypoints prediction which hardly relies on previous data that can be stored in the GRU memory. Moreover, GRU has been proven to be able to eliminate the vanishing gradient problem in a standard recurrent neural network (RNN) using so-called update and reset gates. Besides that, GRU has a better performance-cost ratio than the other RNN layers as it can be trained faster\cite{gru2}. Since there are three waypoints that will be predicted, the process inside the dashed green line box on Figure \ref{fig:netarch} is looped over three times during forward pass. In the first loop, GRU takes the features as an initial hidden state and uses the current waypoint coordinate in BEV space (local vehicle coordinate), route location coordinate transformed to BEV space, and current speed (measured in m/s) as the inputs. To be noted, the initial value for the current waypoint coordinate is the vehicle's local coordinate which is always at (0,0) positioned at the bottom-center of the semantic depth cloud (SDC) as described in Subsection \ref{subs:perception}. Global ($x_{g}$,$y_{g}$) to local ($x_{l}$,$y_{l}$) coordinate transformation can be done with (\ref{eq:bev_transform}).

\begin{equation} \label{eq:bev_transform}
	\begin{bmatrix}
	x_{l} \\
	y_{l} 
	\end{bmatrix} = 
	\begin{bmatrix}
	\cos(90+\theta_v) & -\sin(90+\theta_v) \\
	\sin(90+\theta_v) & \cos(90+\theta_v)
	\end{bmatrix}^T
	\begin{bmatrix}
	x_{g} - x_{vg}\\
	y_{g} - y_{vg}
	\end{bmatrix}
\end{equation}

The relative distance is simply the subtraction between route's global coordinate ($x_{g}$,$y_{g}$) and vehicle's global coordinate ($x_{vg}$,$y_{vg}$). We use $90+\theta_v$ (vehicle rotation degree) in the rotation matrix since the GPS compass is oriented to the north. Then, the next hidden state coming out from the GRU is biased by the prediction of the traffic light state and stop sign which is encoded by a linear layer and simply added through element-wise summation. For waypoints prediction, we use a linear layer to decode the biased hidden state into $\Delta x$ and $\Delta y$. Then, the next waypoint can be obtained with (\ref{eq:wp}).

\begin{equation} \label{eq:wp}
	x_{i+1}, y_{i+1} = (x_{i}+\Delta x), (y_{i}+\Delta y),
\end{equation}

\noindent where $i$ is the current step in the loop process (dashed green line box). On the next loop, the current hidden state (before biased by traffic light state and stop sign) is taken by the GRU to predict the next hidden state with the first waypoint replacing the vehicle's local coordinate (0,0) as the input (along with the same route location and speed measurement). At the end of the looping process, there will be three predicted waypoints and a latent space biased by traffic light state and stop sign prediction. As shown in Figure \ref{fig:netarch}, we use two agents that act as the final decision-makers. The first agent is a multi-layer perceptron (MLP) network composed of two linear layers and a ReLU that decodes the latent space into a set of vehicular controls in a normalized range of 0 to 1 (see Subsection \ref{subs:dataset} for more details). The second agent is two PID controllers (lateral and longitudinal) that compute the predicted waypoints along with the current speed into a set of vehicular controls as summarized in Algorithm \ref{alg:pid}. We set $Kp, Ki, Kd$ parameters for each PID controller as the same with Chen et al.\cite{lbc} and Prakash et al. work\cite{transfuser}. Each agent computes diverse levels of steering, throttle, and brake. Then, the final control actions that actually drive the ego vehicle are made by a control policy defined in Algorithm \ref{alg:control}.

\begin{algorithm}[t] 
	\SetAlgoLined
	$\Theta$ = $\frac{\omega\rho_1+\omega\rho_2}{2}$; $\theta$ = $\tan^{-1}\big(\frac{\Theta[1]}{\Theta[0]}\big)$\\
	$\gamma$ = $2\times||\omega\rho_1 - \omega\rho_2||_F$

	PID steering = Lateral PID($\theta$)\\
	PID throttle = Longitudinal PID($\gamma-\nu$)\\
	

	\caption{PID agent}
	\label{alg:pid}
	\begin{tablenotes}
		\small
		\item --------------------------------------------------------------------------------
		\item $\theta$: heading angle based on first and second waypoints, $\omega\rho_1$ $\omega\rho_2$.
		\item $\gamma$: desired speed, 2 $\times$ Frobenius norm between $\omega\rho_1$ and $\omega\rho_2$.
		\item $\nu$: current speed measured by speedometer.
	\end{tablenotes}
\end{algorithm}

\begin{algorithm}[t] 
	\SetAlgoLined
	
	\uIf{MLP throttle $\geq$ 0.2 and PID throttle $\geq$ 0.2}{
		steering = $\beta_{00}$MLP steering $+$ $\beta_{10}$PID steering\\
		throttle = $\beta_{01}$MLP throttle $+$ $\beta_{11}$PID throttle\\
		brake = 0\\
	}
	\uElseIf{MLP throttle $\geq$ 0.2 and PID throttle $<$ 0.2}{
		steering = MLP steering\\
		throttle = MLP throttle\\
		brake = 0\\
	}
	\uElseIf{MLP throttle $<$ 0.2 and PID throttle $\geq$ 0.2}{
		steering = PID steering\\
		throttle = PID throttle\\
		brake = 0\\
	}
	\uElse{
		steering = 0; throttle = 0; PID brake = 1\\
		brake = $\beta_{02}$MLP brake $+$ $\beta_{12}$PID brake\\
	}
	
	\caption{Control Policy}
	\label{alg:control}
	\begin{tablenotes}
		\small
		\item --------------------------------------------------------------------------------
		\item $\beta\in\{0,...,1\}^{2\times3}$ is a set of control weights initialized with:
		\item $\beta_{00} = \frac{\alpha_4}{\alpha_4+\alpha_7}$; $\beta_{10} = 1 - \beta_{00}$
		\item $\beta_{01} = \frac{\alpha_5}{\alpha_5+\alpha_7}$; $\beta_{11} = 1 - \beta_{01}$ 
		\item $\beta_{02} = \frac{\alpha_6}{\alpha_6+\alpha_7}$; $\beta_{12} = 1 - \beta_{02}$
		\item where $\alpha_4, \alpha_5, \alpha_6, \alpha_7$ are loss weights for steering loss, throttle loss, brake loss, and waypoints loss respectively computed by\\MGN algorithm\cite{mgn} (see Subsection \ref{subs:train} for more details).
		
	\end{tablenotes}
\end{algorithm}

There are two reasons why we predict three waypoints even though only two of them are used by the PID agent. First, the last waypoint prediction ensures that the final hidden state contains the information of the second waypoint which is being used by the GRU as its input. Therefore, it makes the MLP agent act based on the same information as provided to the PID agent. Second, by predicting an extra waypoint, neurons in the GRU and the waypoint prediction layer can have more learning experiences. As described in Algorithm \ref{alg:control}, to make the model becomes more cautious in driving the ego vehicle, we only consider full brake and set the minimum active threshold for the throttle level to 0.2 for each agent.

\section{Experiment Setup}\label{sec:exp_setup}
In this section, we describe the data generation process and data representation. Then, we define the task and explain several scenarios in evaluating the model. Then, we also explain the training configuration including loss function formulation and hyperparameter tuning. Finally, we explain some metrics used to determine model performance.

\subsection{Dataset and Representation}\label{subs:dataset}

We consider imitation learning, especially behavior cloning where the goal is to learn a policy $\pi$ by mimicking the behavior of an expert with a policy $\pi^*$\cite{imitation1}\cite{imitation2}. We define the policy as a mapping function that maps inputs to waypoints, steering, throttle, and brake levels which can be approximated with a supervised learning paradigm. Therefore, we use CARLA simulator\cite{carla} to generate dataset for training and validation. As described in Table \ref{tab:data_info}, we use all available maps and weather preset to create a more varying simulation environment. We also spawn non-player characters (NPC) to mimic real-world conditions. In generating the dataset, an ego vehicle driven by an expert with privileged information is rolled out to retrieve a set of data for every 500ms. One set of data consists of RGB image and depth map along with semantic segmentation ground truth and the corresponding expert trajectory, speed measurement, and vehicular controls. The trajectory is defined by a set of transformed 2D waypoints in bird's eye view (BEV) space (local vehicle coordinate), while the vehicular controls is the record of level of steering, throttle, and brake at the time. For comparison purposes, we also gather LiDAR point clouds which are needed by other models. Following the configuration used by Prakash et al.\cite{transfuser}, we give the expert a set of predefined routes to follow in driving the ego vehicle. Each route is defined with a sequence of GPS coordinates provided by the global planner and high-level navigational command (e.g., turn left, turn right, follow the lane, etc.). There are three kinds of routes sets namely long, short, and tiny. In the long routes set, the expert must drive for 1000-2000 meters comprising around 10 intersections for each. In the short routes set, the expert must drive for 100-500 meters comprising 3 intersections for each. In the tiny routes set, the expert must complete one turn or one go-straight in an intersection or turn. To be noted, the number of routes for each kind of route set is different on each CARLA town depending on the map topography, road length, and other characteristics. Town01 to Town06 have all kinds of route sets, while Town07 and Town10 only have short and tiny route sets. We create two datasets, one for clear noon-only evaluation and one for all weathers evaluation (see Subsection \ref{subs:problem} for more details).

\begin{table}[t] 
	\caption{CARLA Data Generation Setting\cite{transfuser}}
	\begin{center}
		\resizebox{\linewidth}{!}{%
		\begin{tabular}{p{0.275\linewidth}p{0.6\linewidth}}
			\toprule
			Maps & Town01, Town02, Town03, Town04, Town05, Town06, Town07, Town10\\
			\midrule
			Route sets* & Long (1000-2000m)\\
			            & Short (100-500m)\\
			            & Tiny (one turn or one go-straight)\\
			\midrule
			Weather presets & Clear noon, Clear sunset, Cloudy noon, Cloudy sunset, Wet noon, Wet sunset, Mid rainy noon, Mid rainy sunset, Wet cloudy noon, Wet cloudy sunset, Hard rain noon, Hard rain sunset, Soft rain noon, Soft rain sunset\\
			\midrule
			Non-player characters & Other vehicles (truck, car, bicycle, motorbike) and pedestrians\\
			\midrule
			Object classes & 0: Unlabeled, 1: Building, 2: Fence, 3: Other, 4: Pedestrian, 5: Pole, 6: Road lane, 7: Road, 8: Sidewalk, 9: Vegetation, 10: Other vehicles, 11: Wall, 12: Traffic sign, 13: Sky, 14: Ground, 15: Bridge, 16: Rail track, 17: Guard rail, 18: Traffic light, 19: Static object, 20: Dynamic object, 21: Water, 22: Terrain\\
			\midrule
			CARLA version & 0.9.10.1\\
			\bottomrule                             
		\end{tabular}
	}
	\end{center}
	\label{tab:data_info}
	\begin{tablenotes}\small
		\item *The number of routes in long, short, and tiny route sets is different in each town due to varying map complexity and characteristic.
	\end{tablenotes}
\end{table}

Each generated dataset is expressed as $\mathbb{D} = \{(\mathbb{X}^i,\mathbb{Y}^i)\}^J_{i=1}$ where $J$ is the size of the dataset. $\mathbb{X}$ is considered as a set of inputs composed of RGB image, depth map, LiDAR point clouds, speed measurement, GPS locations, and high-level navigational command at a time. To be noted, our model does not need preprocessed LiDAR point clouds and high-level navigational command to drive the ego vehicle. Meanwhile, $\mathbb{Y}$ is considered as a set of outputs composed of semantic segmentation ground truth, waypoints, and the record of vehicular controls at the time together with the state of traffic light and stop sign appearance for additional supervision. Originally, RGB image and depth map are retrieved at a resolution of $300 \times 400$ then cropped to $256 \times 256$ for some reasons described in Subsection \ref{subs:perception}. Thus, both RGB image and depth map are represented as $\mathbb{R} \in \{0,...,255\}^{3 \times 256 \times 256}$ which is the set of 8-bit value in a form of RGB channel ($C$) $\times$ height ($H$) $\times$ width ($W$). Then, the true depth value for each pixel $i$ in the depth map can be decoded with (\ref{eq:depth}).

\begin{equation} \label{eq:depth}
	\mathbb{R}^{dec}_i = \frac{R_i + 256G_i + 256^2B_i}{256^3-1} \times 1000,
\end{equation}

\noindent where $\mathbb{R}^{dec}_i$ is the decoded true depth of pixel $i$, ($R_i,G_i,B_i$) are stored 8-bit value of pixel $i$, 256 is the highest decimal value of 8-bit, and 1000 is the actual depth range of RGBD camera in meters. Meanwhile, LiDAR point clouds are converted into 2-bin histogram over a 2D BEV image $\mathbb{R}^{2 \times 256 \times 256}$ representing the point above and on/below the ground plane\cite{transfuser}\cite{lidarpreprocessed0}\cite{lidarpreprocessed1}. Then, semantic segmentation ground truth is represented as $\mathbb{R} \in \{0,1\}^{23 \times 256 \times 256}$ where 23 is the number of classes mentioned in Table \ref{tab:data_info} with 0 if the pixel does not belong to the class and 1 if the pixel belongs to the class. Then, the waypoints are represented in BEV space with $\{\omega\rho_i = (x_i,y_i)\}^3_{i=1}$. Keep in mind that the center (0,0) of the BEV space (local vehicle coordinate) is on the ego vehicle itself positioned at the bottom-center. The model estimates the vehicular controls in a normalized range of 0 to 1, then they will be denormalized to their original value with steering $\in\{-1,...,1\}$, throttle $\in\{0,...,0.75\}$, and brake $\in\{0,1\}$. For the traffic light state and stop sign prediction, we set 1 if a red light/stop sign appeared, otherwise, they are 0. Meanwhile, speed measurement (in m/s) and GPS locations are sparse, and high-level navigational commands are one-hot encoded.

\subsection{Task and Scenario} \label{subs:problem}
In this research, we consider the point-to-point navigation task where the model is obligated drive the ego vehicle properly following a set of predefined routes and traffic regulations. Following standard CARLA protocol, the routes are defined in the form of sparse GPS locations given by the global planner. The model drivability is evaluated in a variety of areas with different characteristics (e.g., urban, rural, highway, etc.) and various weathers. The main goal is to complete the routes while safely reacting to any events whether in normal or adversarial situations. For example, the model must avoid a collision with a pedestrian that suddenly crosses the street, or with another vehicle when there is a double green light error at an intersection. To achieve convincing evidence, we consider several scenarios of experiments as follows: 
\begin{itemize}
	\item 1W-N: Clear noon only with normal situations. In this scenario, we train the model on all available maps and route sets excluding Town05 short and tiny sets which are used for validation, and leave Town05 long set (10 long routes) for evaluation purposes. Town05 is chosen as it is large and has a complex characteristic. During evaluation, all Non-Player Characters (NPC) behave normally following the traffic rules. We run the experiment three times and calculate the average performance and its standard deviation. The model is expected to be able to drive the ego vehicle properly by not violating the traffic regulations or any other kind of infraction.
	\item 1W-A: Clear noon only with adversarial situations. This scenario is similar to 1W-N, however, the NPC is behaving abnormally which can cause collisions (e.g., the pedestrian or bicyclist is crossing the street suddenly). Intentionally, we also make the traffic light manager create a state where double green lights appear at an intersection. Thus, the ego vehicle may collide with another vehicle coming from a different path. The purpose of this condition is to mimic the event of an ambulance or firefighter truck skipping the traffic light due to emergency situations. Besides driving the ego vehicle properly, the model is expected to be able to safely react and avoid collisions in these adversarial situations. 
	\item AW-N: All weathers with normal situations. This scenario is similar to 1W-N, however, the model is trained and validated with the dataset in which the weather is changed dynamically. Then, the model is evaluated on Town05 long set one time for each weather preset mentioned in Table \ref{tab:data_info}. Thus, we calculate the average and standard deviation of model performance over fourteen times of running as there is fourteen weather presets. The model is expected to be able to adapt to various conditions. 
	\item AW-A: All weathers with adversarial situations. This scenario is similar to AW-N but with adversarial situations as described in 1W-A. This is the hardest scenario in our experiment where model performance is justified based on the robustness against various weather conditions and the ability of responding to adversarial situations. 
\end{itemize}


\subsection{Training Configuration}\label{subs:train}
To learn multiple tasks simultaneously in an end-to-end manner, several loss functions need to be defined first. For semantic segmentation loss function ($\mathcal{L}_{SEG}$), we use a combination of binary cross-entropy and dice loss that can be calculated with (\ref{eq:bcedice}). With this formulation, we can obtain the advantage of distribution-based and region-based losses at the same time\cite{mgn}\cite{unetblock}. Giving extra loss criterion to the semantic segmentation task is necessary as the rest of the network architecture depends on it. 

\begin{equation} \label{eq:bcedice}
	\begin{split}
		\mathcal{L}_{SEG} & = \bigg( \frac{1}{N} \sum_{i = 1}^{N} y_i log(\hat{y}_i) + (1-y_i) log(1-\hat{y}_i) \bigg)\\
		& + \bigg( 1 - \frac{2|\hat{y} \cap y|}{|\hat{y}| + |y|} \bigg),
	\end{split}
\end{equation}

\noindent where $N$ is the number of pixel elements at the output layer of the semantic segmentation decoder. Then, $y_i$ and $\hat{y}_i$ are the value of $i^{th}$ element of the ground truth $y$ and prediction $\hat{y}$ respectively. Meanwhile, we use a simple L1 loss as formulated in (\ref{eq:l1loss}) for the other tasks: traffic light state loss ($\mathcal{L}_{TL}$), stop sign loss ($\mathcal{L}_{SS}$), steering loss ($\mathcal{L}_{ST}$), throttle loss ($\mathcal{L}_{TH}$), brake loss ($\mathcal{L}_{BR}$), and waypoints loss ($\mathcal{L}_{WP}$).

\begin{equation} \label{eq:l1loss}
	\mathcal{L}_{\{TL,SS,ST,TH,BR,WP\}} = |\hat{y} - y|
\end{equation}

To be noted, only $\mathcal{L}_{WP}$ needs to be averaged as there are three predicted waypoints. As explained in Subsection \ref{subs:controller}, the model outputs $\Delta x$ and $\Delta y$ (gap in meters) instead the exact x,y-coordinate location directly. Thus, the waypoints need to be calculated first with (\ref{eq:wp}) before computing the loss. Meanwhile, the prediction of vehicular controls (steering, throttle, brake) need to be denormalized first as described in Subsection \ref{subs:dataset}. Finally, the total loss that is covering all tasks can be computed with (\ref{eq:totalloss}).

\begin{equation} \label{eq:totalloss}
	\begin{split}
		\mathcal{L}_{TOTAL} & = \alpha_1\mathcal{L}_{SEG} + \alpha_2\mathcal{L}_{TL} + \alpha_3\mathcal{L}_{SS}\\
		& + \alpha_4\mathcal{L}_{ST} + \alpha_5\mathcal{L}_{TH} + \alpha_6\mathcal{L}_{BR} + \alpha_7\mathcal{L}_{WP},
	\end{split}
\end{equation}

\begin{table*}
	\caption{Model Specifications}
	\begin{center}
		\resizebox{\textwidth}{!}{%
			\begin{tabular}{ccccc}
				\toprule
				Model & Total Parameters$\downarrow$ & GPU Memory Usage $\downarrow$ & Model Size $\downarrow$ & Input/Sensor\\
				\toprule
				CILRS & 12693331 & 2143 MB & 50.871 MB & RGB camera, speedometer, high-level command\\
				AIM & 21470722 & 2217 MB & 86.033 MB & RGB camera, GPS, speedometer\\
				LF & 32644098 & 2303 MB & 130.808 MB & RGB camera, LiDAR, GPS, speedometer\\
				GF & 40919554 & 2367 MB & 163.944 MB & RGB camera, LiDAR, GPS, speedometer\\
				TF &  66293634 & 2553 MB & 265.617 MB & RGB camera, LiDAR, GPS, speedometer\\
				Ours & 20985934 & 2197 MB & 84.984 MB & RGBD camera, GPS, speedometer\\
				\bottomrule                             
			\end{tabular}
		}
	\end{center}
	\label{tab:model_compare}
	\begin{tablenotes}\small
		\item GPU memory usage is measured by NVIDIA GeForce RTX 3090 while model size is measured based on Ubuntu 20 system. We assume that models with fewer trainable parameters (number of neurons) and less GPU utilization will inference faster. We cannot compute the inference speed fairly since we run multiple parallel experiments at the same time so that the GPU computation performance becomes very fluctuative. CILRS: Conditional Imitation Learning-based model\cite{cilrs} (R: using ResNet\cite{resnet}, S: with Speed input), AIM: Auto-regressive IMage-based model\cite{transfuser}, LF: Late Fusion-based model\cite{transfuser}, GF: Geometric Fusion-based model\cite{geo_fusion}\cite{geo_fusion1}\cite{geo_fusion2}, TF: TransFuser model\cite{transfuser}. 
	\end{tablenotes}
\end{table*}

\noindent where $\alpha_{1,..,7}$ is the loss weight for each task. We use an adaptive loss weighting algorithm called modified gradient normalization (MGN) to tune the loss weights adaptively for each training epoch\cite{mgn}. For a multi-task model, balancing task learning by modifying the gradient signal is necessary to prevent imbalance problems where the model tends to focus on a certain task only. We use Adam optimizer with decoupled weight decay of 0.001 to train the model until convergence\cite{adamw}\cite{optim_adam}. The initial learning rate is set to 0.0001 and reduced gradually by half if there is no drop on validation $\mathcal{L}_{TOTAL}$ in 3 epochs in a row. Additionally, the training will be stopped if there is no improvement in 15 epochs in a row to prevent unnecessary computational costs. We implement our model with PyTorch framework\cite{torch} and train it on NVIDIA GeForce RTX 3090 with a batch size of 20.

\subsection{Evaluation Metrics}\label{subs:eval}
In the evaluation process, there are several metrics that can be used to justify model performance in several aspects of driving. Following CARLA leaderboard evaluation setting\footnote{\url{https://leaderboard.carla.org}}, we use driving score (DS) as the main metric where the higher the driving score means the better the model. The driving score can be computed with (\ref{eq:ds}).

\begin{equation} \label{eq:ds}
	DS = \frac{1}{N_r} \sum_{i = 1}^{N_r} RC_{i}IP_i
\end{equation}

The DS for the $i^{th}$ route ($DS_i$) is a simple multiplication between the percentage of route completion of route $i$ ($RC_i$) and the infraction penalty of route $i$ ($IP_i$). Then, the final driving score can be calculated by averaging over $N_r$, the number of routes in the route set. $RC_i$ can be simply obtained by dividing the completed distance of route $i$ with the total length of route $i$. However, if the ego vehicle drives offroad (e.g., drives on the sidewalk), then the path where the ego vehicle drives offroad is not counted and yields a reduced $RC_i$. Meanwhile, $IP_i$ can be computed with (\ref{eq:ip}). 

\begin{equation} \label{eq:ip}
	IP_i = \prod_j^{M} ({p_i^j})^{\text{\#infractions}_j} ,
\end{equation}


\noindent where $M$ is the set of infraction types considered for the evaluation process. The $IP_i$ for each model start with an ideal 1.0 base score, which is reduced as an infraction is committed. Ordered by its severity, we consider the following types of infraction $M$ and penalty values $p^j$:
\begin{itemize}
	\item Collision with pedestrians: 0.50
	\item Collision with other vehicles: 0.60
	\item Collision with others (static elements): 0.65
	\item Red light violation: 0.70
	\item Stop sign violation: 0.80
\end{itemize}

The final RC and IP scores can be obtained by averaging over $N_r$ as similar to the final DS calculation. To save computation costs, the evaluation process on route $i$ will be stopped if the ego vehicle deviates more than 30 meters from the assigned route or does not take any actions for 180 seconds. Then, the evaluation process continues to the next route for further performance calculation. 

\section{Result and Discussion}\label{sec:result}

\begin{table*}
	\caption{Performance Comparison in All Scenarios Evaluation}
	\begin{center}
		\resizebox{\textwidth}{!}{%
			\begin{tabular}{ccccccccccc}
				\toprule
				\multirow{2}{*}{Scenario} & \multirow{2}{*}{Model} & \multirow{2}{*}{DS$\uparrow$} & \multirow{2}{*}{RC$\uparrow$} & \multirow{2}{*}{IP$\uparrow$} & \multicolumn{3}{c}{Collision$\downarrow$} & \multicolumn{2}{c}{Violation$\downarrow$}&\multirow{2}{*}{Offroad$\downarrow$}\\
				&             &                  &       &     &Pedestrian&Vehicle&Others&Red&Stop&\\
				\toprule 
				& CILRS    & 8.291 $\pm$0.571 & 11.149&0.819&0.000&1.060&0.019&0.207&0.000&1.042\\
				& AIM      &41.191 $\pm$1.047 & 92.794&0.456&0.000&0.043&0.004&0.108&0.088&0.014\\
				Normal                & LF       &42.770 $\pm$2.334 & 73.836&0.457&0.000&0.032&0.004&0.043&0.072&0.055\\
				Clear Noon            & GF       &37.320 $\pm$7.150 & 82.592&0.480&0.000&0.056&0.000&0.089&0.123&0.052\\
				(1W-N)$^{\dagger}$    & TF       &35.273 $\pm$1.115 & 57.680&0.710&0.000&0.020&0.000&0.097&0.037&0.085\\
				&{\bf Ours}&{\bf 48.428 $\pm$1.467} & 84.270&0.625&0.000&0.038&0.000&0.048&0.075&0.011\\
				\hdashline\noalign{\vskip 0.75ex}
				& Expert   &73.553 $\pm$5.086 &100.000&0.735&0.000&0.051&0.000&0.019&0.000&0.000\\
				\midrule
				& CILRS    & 8.261 $\pm$0.924 & 13.237&0.657&0.119&0.512&0.095&0.258&0.000&1.012\\
				& AIM      &25.541 $\pm$2.257 & 79.141&0.412&0.018&0.113&0.005&0.175&0.041&0.081\\
				Adversarial           & LF       &35.310 $\pm$2.510 & 58.744&0.658&0.012&0.049&0.000&0.102&0.018&0.031\\
				Clear Noon            & GF       &32.423 $\pm$1.042 & 65.291&0.602&0.012&0.148&0.000&0.049&0.023&0.094\\
				(1W-A)$^{\dagger}$    & TF       &24.740 $\pm$0.948 & 37.921&0.788&0.016&0.074&0.008&0.121&0.008&0.101\\
				&{\bf Ours}&{\bf 35.982 $\pm$2.105} & 76.189&0.476&0.034&0.062&0.000&0.214&0.020&0.014\\
				\hdashline\noalign{\vskip 0.75ex}
				& Expert   &41.579 $\pm$1.576 & 68.225&0.696&0.000&0.086&0.000&0.081&0.000&0.000\\
				\midrule
				& CILRS    & 7.376 $\pm$0.996 &  9.569&0.869&0.000&0.849&0.020&0.071&0.000&0.964\\
				& AIM      &39.613 $\pm$3.644 & 84.467&0.542&0.000&0.022&0.018&0.090&0.076&0.030\\
				Normal                & LF       &36.890 $\pm$7.067 & 48.747&0.805&0.000&0.063&0.001&0.023&0.050&0.041\\
				All Weathers          & GF       &20.446 $\pm$5.281 & 29.616&0.831&0.000&0.051&0.000&0.079&0.022&0.063\\
				(AW-N)*              & TF       &16.672 $\pm$4.175 & 26.425&0.843&0.000&0.011&0.014&0.082&0.010&0.427\\
				&{\bf Ours}&{\bf 47.133 $\pm$5.276} & 77.427&0.653&0.000&0.069&0.021&0.031&0.056&0.080\\
				\hdashline\noalign{\vskip 0.75ex}
				& Expert   &75.815 $\pm$5.045 &100.000&0.758&0.000&0.058&0.000&0.012&0.000&0.000\\
				\midrule
				& CILRS    & 5.234 $\pm$0.869 &  8.663&0.778&0.119&0.039&0.148&0.357&0.000&1.049\\
				& AIM      &30.207 $\pm$5.857 & 76.399&0.471&0.013&0.066&0.028&0.147&0.041&0.047\\
				Adversarial           & LF       &22.592 $\pm$7.039 & 32.525&0.808&0.371&0.088&0.010&0.050&0.021&0.069\\
				All Weathers          & GF       &14.799 $\pm$3.860 & 20.113&0.865&0.007&0.120&0.007&0.036&0.008&0.069\\
				(AW-A)*              & TF       &11.167 $\pm$2.434 & 20.965&0.808&0.003&0.054&0.033&0.124&0.011&0.273\\
				&{\bf Ours}&{\bf 31.055 $\pm$2.700} & 64.132&0.531&0.031&0.128&0.049&0.106&0.022&0.134\\
				\hdashline\noalign{\vskip 0.75ex}
				& Expert   &39.475 $\pm$6.792 & 71.474&0.643&0.002&0.108&0.001&0.078&0.000&0.000\\
				\bottomrule
			\end{tabular}
		}
	\end{center}
	\label{tab:8T}
	\begin{tablenotes}\small
		\item The best performance is defined by the highest driving score (DS) on each scenario. CILRS: Conditional Imitation Learning-based model\cite{cilrs} (R: using ResNet\cite{resnet}, S: with Speed input), AIM:  Auto-regressive IMage-based model\cite{transfuser}, LF: Late Fusion-based model\cite{transfuser}, GF: Geometric Fusion-based model\cite{geo_fusion}\cite{geo_fusion1}\cite{geo_fusion2}, TF: TransFuser model\cite{transfuser}. 
		\item $^{\dagger}$The result is averaged over three times of experiment run.
		\item *The result is averaged over fourteen times of experiment run with one time running for each weather preset mentioned in Table \ref{tab:data_info}.
	\end{tablenotes}
\end{table*}

As mentioned in Subsection \ref{subs:problem}, we evaluate the model in several scenarios to understand multiple aspects of driving. For a comparative study, we pick CILRS\cite{cilrs} (Conditional Imitation Learning using ResNet\cite{resnet} and Speed input) as the representative of a model that needs high-level navigational commands to drive the ego vehicle. Then, we replicate several models developed by Prakash et al.\cite{transfuser} as the representative of a model that does not need high-level navigational commands to drive the ego vehicle. To be more detailed, we replicate four models namely AIM (Auto-regressive IMage), LF (Late Fusion), GF (Geometric Fusion), and TF (TransFuser). These models have the same module that is responsible for determining vehicular controls, however, their perception modules are completely different from one another. AIM only uses an RGB camera as the main source of information for its perception module. Meanwhile, the other models combine a front RGB camera and LiDAR sensor but with different fusion strategies. TF uses transformers to learn the relationship between two unique features. GF use geometric transformation inspired by Liang et al.\cite{geo_fusion}\cite{geo_fusion1}\cite{geo_fusion2} for fusing the extracted features. Meanwhile, LF only uses element summation to fuse both features and let the next layer learn its correlation. To ensure that the comparison is conducted fairly, we use the same camera settings for all models and consider an ROI of $256 \times 256$ described in Prakash et al.'s works\cite{transfuser}. The model specification details can be seen in Table \ref{tab:model_compare}. Furthermore, we conduct an ablation study by modifying some modules in the model architecture and changing the control policy to understand their contribution.

\begin{figure*}
	\centering
	\includegraphics[width=\textwidth]{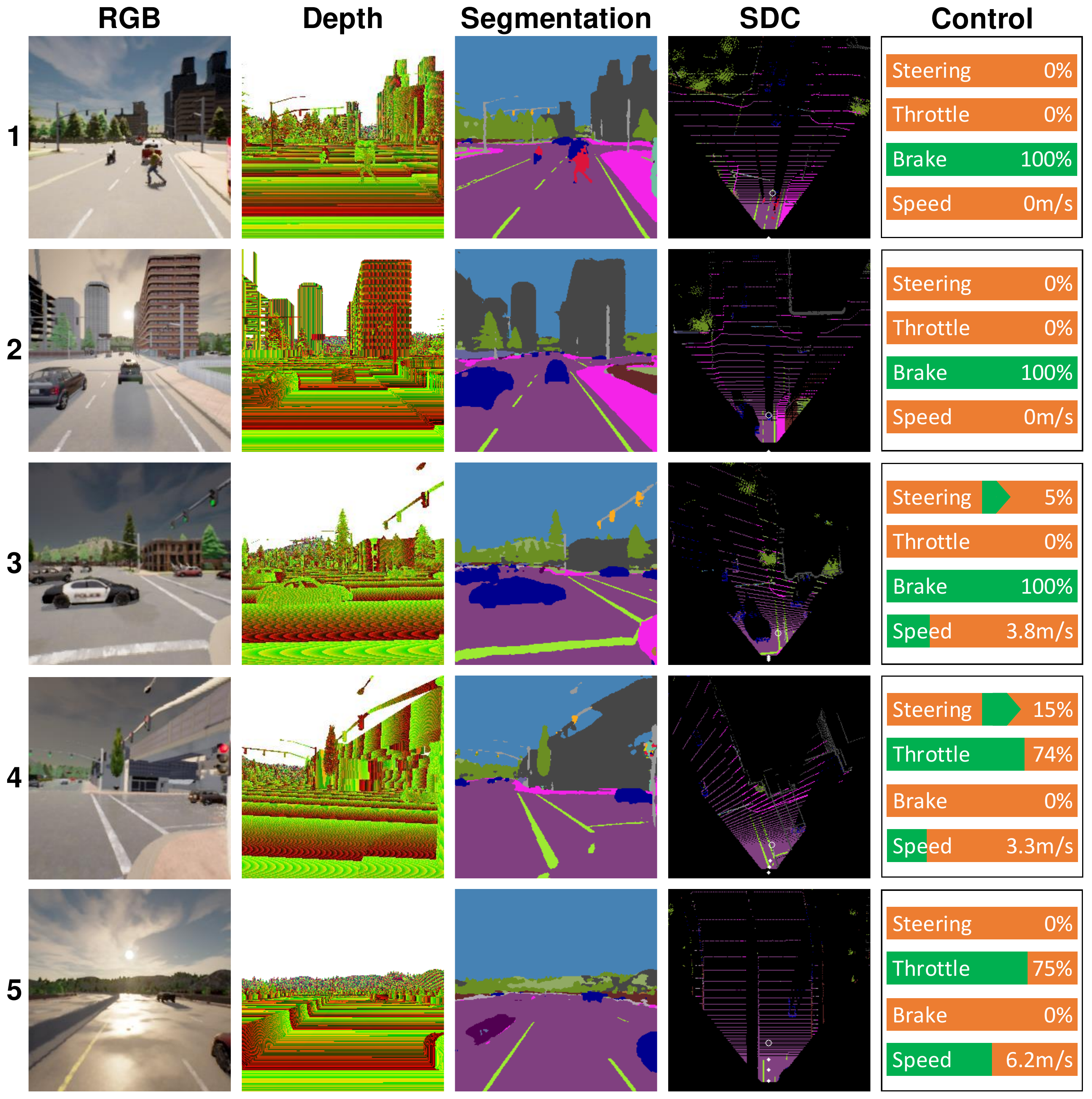}
	\caption{Qualitative result of our model performance.} 
	\label{fig:inference}
	\begin{tablenotes}\small
		\item 1. Clear Noon: The model manages to stop the ego vehicle as a pedestrian is crossing suddenly. Therefore, the pedestrian is not getting hit by the vehicle.
		\item 2. Cloudy Sunset: Even though there are some vacant spaces in front of the ego vehicle, the model tends to stop behind any vehicles shown on the front camera at the intersection. The model is doing behavior cloning to the expert which is doing the same thing.
		\item 3. Mid Rainy Noon: The model reacts properly and avoids collision by doing instant braking when there is a vehicle driving ahead due to double green light error at the intersection.
		\item 4. Hard Rain Sunset: After the traffic light turns green, the model drives to the right getting close to the given route (white hollow circle inside semantic depth cloud (SDC)). However, it drives offroad as the sidewalk is misclasified as road.
		\item 5. Wet Sunset: The model drives fast on a highway. Although it cannot segment the road barrier properly, the model manages to stay on the lane thanks to the well-segmented road lane and the information on vacant regions in the SDC map.
	\end{tablenotes}
\end{figure*}

To be noted, the main evaluation metric used to determine the best model is the driving score (DS) which is the multiplication of route completion (RC) and infraction penalty (IP). Keep in mind that having a higher RC or IP does not mean that the model is better. The model may have a higher RC by disobeying the traffic rules so it can keep going to achieve further route distance but drive terribly bad. The model may also have small infractions counts (result in higher IP) due to the low percentage of the completed route where any collisions or traffic violations can happen if it travels further. Therefore, the most appropriate metric is DS as it combines both aspects of driving in RC and IP: drive as far as possible with small infractions as little as possible. As mentioned in Subsection \ref{subs:problem}, we calculate the average and the standard deviation of each model performance over three times of experiment run in clear noon-only evaluation (1W-N and 1W-A) and over fourteen times of experiment run in all weathers evaluation (AW-N and AW-A). The evaluation score can be seen in Table \ref{tab:8T}. In addition, we add several driving footages on various weather conditions as shown in Figure \ref{fig:inference}.

To reflect the intuitive performance on each task independently, we also conduct an inference test on the expert driving data and do a comparative study with some recent task-specific models. To be more detailed, we compare our model against CILRS for vehicular controls (steering, throttle, brake) estimation task. Then, we compare our model against AIM, LF, GF, and TF for the waypoints prediction task. Meanwhile, for semantic segmentation comparison, we train and evaluate DeepLabV3+ segmentation model\cite{deeplabv3plus} with ResNet101 backbone\cite{resnet} pre-trained on ImageNet\cite{imagenet}. Finally, for the performance comparison of traffic light state and stop sign prediction, we train and evaluate a classifier model based on Efficient Net B7\cite{effnet} pre-trained on ImageNet\cite{imagenet}.

\subsection{Drivability in Normal and Adversarial Situations}\label{subs:modeldrivability}
Normally, all road users, including pedestrians, must obey the traffic regulations so that any kind of accident can be prevented. However, this condition is nearly impossible in the real world, especially in a crowded urban area where plenty of vehicles and pedestrians are moving around. No one can guarantee that the pedestrian will always walk only on the sidewalk or not cross the street suddenly. Moreover, there is also a possibility that the traffic light can err (e.g., double green light) due to a certain cause. Therefore, in this research, we evaluate all models in both normal situations (where everything is working as it should be) and adversarial situations (where unexpected events can occur at any time). Specifically, in adversarial situations, we make abnormal events by spawning some pedestrians to cross the street suddenly and making the traffic light error when the ego vehicle is approaching. These events are used to mess up the traffic on purpose. In these situations, the model is expected to be able to react safely and avoid collisions with pedestrians or other vehicles.

Based on Table \ref{tab:8T}, our proposed model has the best drivability by achieving the highest driving scores (DS) in all scenarios. Moreover, our model can be said to be more stable compared to the second-best model, LF, as it achieves a smaller standard deviation of DS in the clear noon-only evaluation (1W-N and 1W-A). Meanwhile, in all weathers evaluation, our model stability is comparable to AIM as the standard deviation is larger in AW-N but smaller in AW-A. This is caused by the variety of weather conditions that challenge the capability of the perception module of each model in extracting stable features for the controller module. Based on the comparison of DS in normal situations and adversarial situations, all models are suffering from such unexpected events. The driving scores achieved by each model in 1W-A and AW-A scenarios are lower than in 1W-N and AW-N scenarios. The performance drop is as expected due to abnormalities that occurred during driving. However, even with these adversarial situations, our model still manages to achieve the best performance meaning that it can react properly to prevent such kinds of infractions.

\subsection{Adaptability to Various Weather Conditions}\label{subs:modeladaptability}
To evaluate all models further, we also change the weather condition in fourteen weather presets provided in CARLA simulator. The purpose of this test is to check the adaptability of the model when deployed in various conditions. In the real world, it is obvious that we cannot expect that the weather will always be sunny. Therefore, conducting adaptability tests in various weather conditions is necessary. Unlike in the clear noon-only test where each model is running three times, we run each model only once for each weather. Then, the average and standard deviation are calculated over fourteen set of scores as described in Subsection \ref{subs:problem}. 

As shown in Table \ref{tab:8T}, all models get their driving score dropped in AW-N and AW-A scenarios compared to their achievement in 1W-N and 1W-A scenarios. Besides that, most of the models have a larger standard deviation on their scores. However, our model still manages to achieve the best performance. The phenomena of performance drop and less stability mean that performing scene understanding to perceive the surrounding is much harder in various weather conditions. The only outlier of this result is the fact that the AIM performance on the adversarial situation in all weathers evaluation is better than in clear noon-only evaluation where its DS is getting improved from 25.541 to 30.207. An answer to this phenomenon lies in the network architecture of AIM. As mentioned in Table \ref{tab:model_compare}, AIM uses front RGB image as the only input for its perception module. Therefore, in the various weather conditions, its capability in extracting informative and stable features can be drastically improved as the rest of the architecture only relies on it.

\begin{table*}
	\caption{Ablation Study in Clear Noon Evaluation}
	\begin{center}
		\resizebox{\textwidth}{!}{%
			\begin{tabular}{ccccccccccc}
				\toprule
				\multirow{2}{*}{Scenario} & \multirow{2}{*}{Policy} & \multirow{2}{*}{DS$\uparrow$} & \multirow{2}{*}{RC$\uparrow$} & \multirow{2}{*}{IP$\uparrow$} & \multicolumn{3}{c}{Collision$\downarrow$} & \multicolumn{2}{c}{Violation$\downarrow$}&\multirow{2}{*}{Offroad$\downarrow$}\\
				&             &                  &       &     &Pedestrian&Vehicle&Others&Red&Stop&\\
				\toprule 
				&No SDC       &39.579 $\pm$10.251& 63.276&0.715&0.000&0.028&0.022&0.062&0.068&0.029\\
				&{\bf Proposed}&{\bf 48.428 $\pm$1.467} & 84.270&0.625&0.000&0.038&0.000&0.048&0.075&0.011\\
				Normal        &MLP          &43.210 $\pm$3.689 & 80.794&0.589&0.000&0.047&0.000&0.078&0.054&0.020\\
				(1W-N)        &PID          &46.351 $\pm$7.879 & 65.673&0.761&0.000&0.022&0.000&0.081&0.044&0.034\\
				&Both         &42.927 $\pm$9.392 & 67.373&0.705&0.000&0.026&0.000&0.072&0.037&0.010\\
				\hdashline\noalign{\vskip 0.75ex}
				&Expert       &71.440 $\pm$5.855 &100.000&0.714&0.000&0.061&0.000&0.011&0.000&0.000\\
				\midrule
				&No SDC       &29.624 $\pm$7.006 & 56.430&0.606&0.030&0.043&0.006&0.256&0.026&0.013\\
				&{\bf Proposed}&{\bf 35.982 $\pm$2.105} & 76.189&0.476&0.034&0.062&0.000&0.214&0.020&0.014\\
				Adversarial   &MLP          &26.936 $\pm$3.773 & 59.178&0.511&0.090&0.109&0.000&0.207&0.015&0.004\\
				(1W-A)        &PID          &32.282 $\pm$1.899 & 58.748&0.635&0.004&0.112&0.010&0.181&0.026&0.017\\
				&Both         &32.396 $\pm$4.531 & 58.367&0.622&0.004&0.108&0.000&0.227&0.012&0.028\\
				\hdashline\noalign{\vskip 0.75ex}
				&Expert       &41.579 $\pm$1.576 & 68.225&0.696&0.000&0.086&0.000&0.081&0.000&0.000\\
				\bottomrule
			\end{tabular}
		}
	\end{center}
\label{tab:ablation}
	\begin{tablenotes}\small
		\item The result is averaged over three times of experiment run. The best performance is defined by the highest driving score (DS) on each scenario. Proposed: the configuration as described in Section \ref{sec:model}, No SDC: there is no semantic depth cloud mapping so that the controller module only receives RGB features, MLP: the model only uses MLP agent on its controller module, PID: the model only uses PID agent on its controller module, Both: the model drives the vehicle if and only if both MLP and PID throttles are above the threshold of 0.2.
	\end{tablenotes}
\end{table*}

\subsection{Models Behavior}\label{subs:modelbehavior}

The CILRS model which takes high-level navigational commands is completely far behind the other models. The problem with the navigational command-based model is the imbalance distribution of discrete navigational commands. It is obvious that the command 'follow lane' or 'go straight' are more than 'turn left' or 'turn right'. As a result, the model tends to fail to turn properly at the intersection. Judging from its low percentage of RC, the model cannot drive any further due to a collision with a pedestrian, another vehicle, or other objects. The model may have failed at the first or second intersection in the given route.

The AIM model has the highest RC, however, it cannot achieve the highest DS as it makes many infractions resulting in the lowest IP in all scenarios. The problem with AIM is it only uses the front RGB image as the only input for its perception module. The reason AIM can travel further compared to the other models in all scenarios is due to its less awareness caused by limited information about the surrounding. AIM tends to just keep going and makes a lot of infractions since the only thing it knows about is the information in front of the ego vehicle. This kind of driving behavior is resulting in the highest RC but with the smallest IP (many infractions) in each scenario.

The RGB-LiDAR fusion-based models (LF, GF, and TF) are too careful which results in smaller RC and higher IP (fewer infractions). During the evaluation process, these models tend to wait for another vehicle to make the first move in an intersection so that they can follow those vehicles. Unlike AIM, these models focus on capturing the surrounding vehicle situation rather than recognizing the traffic light state to decide whether to drive the vehicle or not. Therefore, if there are no other vehicles around, then these models are not going to drive the vehicle any further. In this case, these models may have a better understanding of the surrounding area, however, the extracted features provided by the LiDAR encoder are biasing too much and affecting its ability to recognize the state of the traffic light which cannot be captured by LiDAR.

In our model, the information provided by LiDAR is replaced by the semantic depth cloud (SDC) mapping function. In addition, we use a specific module to recognize the traffic light state and the appearance of stop signs that may disappear during GRU looping. Furthermore, we also use two agents to create more varying driving options. The first agent is the MLP network that decodes the final hidden state given by the GRU. Meanwhile, the second agent is two PID controllers that translate predicted waypoints into vehicular controls. As a result, the model can maintain the trade-off between RC and IP which is resulting in the highest driving score as shown in Table \ref{tab:8T}. Moreover, our model is more efficient compared to LF and AIM as it has fewer parameters and smaller GPU memory usage as described in Table \ref{tab:model_compare}.

\subsection{Ablation Study}\label{subs:ablation}
We conduct ablation study on the clear noon-only evaluation by removing the semantic depth cloud (SDC) mapping function and changing the control policy to understand how the SDC and multi-agent can improve model performance. 

\subsubsection{The Importance of SDC} \label{subss:ablation1}
To understand how important the SDC mapping is, we train and evaluate another model that does not have this function with the same experiment settings for a fair comparison. As shown in Table \ref{tab:ablation}, the 'no SDC' model achieves a lower driving score with a higher standard deviation. This result is as expected since the model loses its capability in performing robust scene understanding. Unlike LiDAR which provides the height information, the SDC holds semantic information for each class on each layer. With this representation, the model can understand many useful information easily. Moreover, by doing concatenation rather than element summation on RGB and SDC features, the fusion block can learn the relation by itself to prevent information loss.

\subsubsection{The Role of Multi-Agent} \label{subss:ablation2}
By default ('Proposed' control policy), the model uses both MLP and PID agents in driving the ego vehicle as described in Subsection \ref{subs:controller}. The model will stop the vehicle if and only if both throttle values are below the threshold of 0.2. If only the MLP agent in which the throttle value is higher than 0.2, then the vehicle is fully controlled by the MLP agent, and so does for the PID agent. By using the same model, we create three more control policies named 'MLP', 'PID', and 'Both'. In 'MLP' and 'PID' control policies, the vehicle is controlled by one agent only, MLP or PID. Meanwhile, the 'Both' control policy uses both agents as similar to the 'Proposed' control policy. However, the model will drive the vehicle if and only if both throttle values are higher than 0.2. This means that if one of the throttle values is below 0.2, the vehicle will stop.

Based on Table \ref{tab:ablation}, the 'Proposed' control policy still achieves the best performance compared to the other policies. With more driving options provided by two agents that represent different aspects of driving, the model can drive the vehicle farther and make a better trade-off between route completion (RC) percentage and infraction penalty (IP). In normal situations, the 'Both' policy achieves the lowest score as it is too careful by considering both decisions made by MLP and PID agents. However, this carefulness can make the 'Both' policy surpasses 'MLP' and 'PID' policies in adversarial situations since being more careful is preferable in handling unexpected abnormal events such as pedestrian crossing suddenly. Meanwhile, the 'MLP' policy achieves the second-best in RC which means that the MLP agent is careless compared to the PID agent and results in a lower IP (many infractions). On the other hand, the PID agent can be said to be better than the MLP agent as it achieves higher DS. This means that the 'PID' policy is better than the 'MLP' policy in managing the trade-off between RC and IP.

\subsection{Task-wise Evaluation} \label{subs:mtlperformance}
The purpose of this evaluation is to analyze model performance in handling multiple perception and control tasks simultaneously. Hence, we conduct a comparative study with some task-specific models to reflect the intuitive performance on each task independently. In this experiment, we configure all models to perform inference on the expert's driving data on the evaluation routes (Town05 long route set) in four different scenarios (1W-N, 1W-A, AW-N, AW-A) which are completely unseen in the training and validation datasets used for imitation learning process. We consider expert's trajectory and vehicular controls (steering, throttle, brake) record along with segmentation map, traffic light state, and stop sign appearance provided by CARLA as the ground truth for evaluation. 


For metric scoring, we use intersection over union (\ref{eq:iou}) to determine the performance on semantic segmentation task.

\begin{equation} \label{eq:iou}
IoU_{SEG} = \frac{|\hat{y} \cap y|}{|\hat{y} \cup y|},
\end{equation}

\noindent where $\hat{y}$ and $y$ are the prediction and ground truth respectively. Meanwhile, for the traffic light state and stop sign predictions, we use a simple accuracy scoring (\ref{eq:acc}) as the metric function.

\begin{equation} \label{eq:acc}
Acc._{\{TL,SS\}} = \frac{TP+TN}{TP+TN+FP+FN},
\end{equation}
 
\noindent where $TP$, $TN$, $FP$, and $FN$ are the true positive, true negative, false positive, and false negative predictions made by the model. Then, to justify model performance in predicting waypoints and vehicular controls, we use mean absolute error (MAE) as the metric function which is the same function used for their loss calculation as mentioned in Subsection \ref{subs:train} (referred as L1 loss). As similar to Table \ref{tab:8T}, the results shown in Table \ref{tab:seg_compare}, \ref{tab:tlss_compare}, \ref{tab:wp_compare}, and \ref{tab:control_compare} are averaged over three inference results for clear noon-only scenarios (1W-N and 1W-A) and averaged over fourteen inference results for all weathers scenarios (AW-N and AW-A).





\subsubsection{Semantic Segmentation} \label{subss:seg_compare}
Based on Table \ref{tab:seg_compare}, it can be said that performing semantic segmentation in varying weather conditions is harder than in a single weather condition. In the varying weather scenarios (AW-N and AW-A), both DeepLabV3+ and our model have lower IoU scores with larger standard deviation than in the clear noon-only scenarios (1W-N and 1W-A). This result is in line with the slight degradation of the driving score shown in Table \ref{tab:8T} and discussed in Subsection \ref{subs:modeladaptability} meaning that performing prediction in various weathers is more challenging. In all scenarios, it is expected that DeepLabV3+ has higher IoU scores with a smaller standard deviation meaning that it has better performance and stability than our model. However, with just a small gap difference in IoU score, our model is still preferable considering the large number of neurons used in DeepLabV3+ architecture that can cause a huge computational load.

\begin{table}[t]
	\caption{Semantic Segmentation Score}
	\begin{center}
		\begin{tabular}{ccc}
			\toprule
			Scenario&Model&$IoU_{SEG}\uparrow$\\
			\toprule 
			\multirow{2}{*}{1W-N}&{\bf DLV3+}&{\bf 0.888 $\pm$0.001}\\
			&Ours&0.883 $\pm$0.001\\
			\midrule
			\multirow{2}{*}{1W-A}&{\bf DLV3+}&{\bf 0.885 $\pm$0.002}\\
			&Ours&0.880 $\pm$0.003\\
			\midrule
			\multirow{2}{*}{AW-N}&{\bf DLV3+}&{\bf 0.884 $\pm$0.003}\\
			&Ours&0.878 $\pm$0.004\\
			\midrule
			\multirow{2}{*}{AW-A}&{\bf DLV3+}&{\bf 0.883 $\pm$0.003}\\
			&Ours&0.879 $\pm$0.004\\
			\bottomrule
		\end{tabular}
	\end{center}
	\label{tab:seg_compare}
	\begin{tablenotes}\small
		\item $IoU_{SEG}$: intersection over union score of semantic segmentation.
		\item DLV3+: DeepLabV3+ segmentation model\cite{deeplabv3plus} with ResNet101 backbone\cite{resnet}.
	\end{tablenotes}
\end{table}

\begin{table}[t]
	\caption{TL State and Stop Sign Prediction Score}
	\begin{center}
		\begin{tabular}{cccc}
			\toprule
			Scenario&Model&$Acc._{TL}\uparrow$&$Acc._{SS}\uparrow$\\
			\toprule 
			\multirow{2}{*}{1W-N}&ENB7&0.967 $\pm$0.008&0.995 $\pm<$0.001\\
			&{\bf Ours}&{\bf 0.986 $\pm$0.003}&{\bf 0.996 $\pm<$0.001}\\
			\midrule
			\multirow{2}{*}{1W-A}&ENB7&0.975 $\pm$0.004&{\bf 0.996 $\pm<$0.001}\\
			&{\bf Ours}&{\bf 0.982 $\pm$0.006}&{\bf 0.996 $\pm<$0.001}\\
			\midrule
			\multirow{2}{*}{AW-N}&ENB7&0.980 $\pm$0.005&{\bf 0.996 $\pm<$0.001}\\
			&{\bf Ours}&{\bf 0.989 $\pm$0.004}&{\bf 0.996 $\pm<$0.001}\\
			\midrule
			\multirow{2}{*}{AW-A}&{\bf ENB7}&{\bf 0.979 $\pm$0.003}&{\bf 0.996 $\pm$0.001}\\
			&Ours&0.979 $\pm$0.004&{\bf 0.996 $\pm$0.001}\\
			\bottomrule
		\end{tabular}
	\end{center}
	\label{tab:tlss_compare}
	\begin{tablenotes}\small
		\item $Acc._{TL}$: accuracy of traffic light (TL) state prediction, $Acc._{SS}$: accuracy of stop sign prediction.
		\item ENB7: A classifier model based on Efficient Net B7 variant\cite{effnet}.
	\end{tablenotes}
\end{table}

\subsubsection{TL State and Stop Sign Prediction} \label{subss:tlss_compare}
In the task of traffic light (TL) state and stop sign prediction, our model has comparable accuracy scores in all scenarios as shown in Table \ref{tab:tlss_compare}. Unlike in the semantic segmentation task, the model only deals with one specific object which is easier to handle. Moreover, although our model uses a smaller Efficient Net version (Efficient Net B3) as its RGB encoder, it manages to be better than a classifier model based on Efficient Net B7 (the best and the biggest amongst Efficient Net model variants) almost in all scenarios. Our model only loses in the AW-A scenario where it has a slightly larger standard deviation in the traffic light state prediction score. Thanks to the end-to-end and multi-task learning strategy, the model can receive more supervision since its prediction is encoded further and used to bias a certain latent space for other task predictions. 

\begin{table}[t]
	\caption{Waypoints Prediction Score}
	\begin{center}
		\begin{tabular}{ccc}
			\toprule
			Scenario&Model&$MAE_{WP}\downarrow$\\
			\toprule 
			&AIM       &0.233 $\pm$0.017\\
			&LF        &0.209 $\pm$0.011\\
			1W-N&GF    &0.231 $\pm$0.010\\
			&TF        &0.183 $\pm$0.009\\
			&{\bf Ours}&{\bf 0.114 $\pm$0.003}\\
			\midrule
			&AIM       &0.326 $\pm$0.042\\
			&LF        &0.307 $\pm$0.039\\
			1W-A&GF   &0.326 $\pm$0.038\\
			&TF        &0.286 $\pm$0.039\\
			&{\bf Ours}&{\bf 0.166 $\pm$0.023}\\
			\midrule
			&AIM       &0.221 $\pm$0.015\\
			&LF        &0.192 $\pm$0.014\\
			AW-N&GF    &0.196 $\pm$0.014\\
			&TF        &0.186 $\pm$0.009\\
			&{\bf Ours}&{\bf 0.120 $\pm$0.006}\\
			\midrule
			&AIM       &0.292 $\pm$0.028\\
			&LF        &0.294 $\pm$0.036\\
			AW-A&GF    &0.271 $\pm$0.025\\
			&TF        &0.249 $\pm$0.018\\
			&{\bf Ours}&{\bf 0.172 $\pm$0.015}\\
			\bottomrule
		\end{tabular}
	\end{center}
	\label{tab:wp_compare}
	\begin{tablenotes}\small
		\item $MAE_{WP}$: mean absolute error of waypoints prediction.
		\item AIM:  Auto-regressive IMage-based model\cite{transfuser}, LF: Late Fusion-based model\cite{transfuser}, GF: Geometric Fusion-based model\cite{geo_fusion}\cite{geo_fusion1}\cite{geo_fusion2}, TF: TransFuser model\cite{transfuser}.
	\end{tablenotes}
\end{table}

\subsubsection{Waypoints Prediction} \label{subss:wp_compare}
As mentioned earlier in Section \ref{sec:result}, we compare our model with AIM, LF, GF, and TF in the waypoints prediction task. These models predict four waypoints, while our model only predicts three waypoints. Yet, it is still considered a fair comparison since we use MAE that averages the error on all predictions. Based on Table \ref{tab:wp_compare}, our model has the lowest MAE with the smallest standard deviation compared to the other models in all scenarios. Thanks to the SDC mapping which lies on BEV space the same as where the coordinates of the waypoint are located. This means that the SDC is proven to play an important role in giving the model a strong intuition in predicting waypoints as it provides the information of free and occupied regions clearly in BEV space. Besides that, performing waypoints prediction in adversarial driving is harder than in normal driving. In a comparison between the result in 1W-N with 1W-A and AW-N with AW-A, the MAE constantly become larger. Furthermore, if we compare the result in 1W-N with AW-N and 1W-A with AW-A, the MAE also constantly become larger which means that performing waypoints prediction in varying weathers is also harder than in one single weather. This result is in line with the result shown in Table \ref{tab:8T} and discussed in Subsection \ref{subs:modeldrivability} and \ref{subs:modeladaptability} where all models are suffered from adversarial situations and varying weather conditions, yielding a lower driving score.


\begin{table}[t]
	\caption{Vehicular Controls Estimation Score}
	\begin{center}
		\resizebox{\linewidth}{!}{%
			\begin{tabular}{ccccc}
				\toprule
				Scenario&Model&$MAE_{ST}\downarrow$&$MAE_{TH}\downarrow$&$MAE_{BR}\downarrow$\\
				\toprule 
				\multirow{2}{*}{1W-N}&{\bf CILRS}&{\bf 0.022 $\pm$0.002}&{\bf 0.052 $\pm$0.002}&0.053 $\pm$0.003\\
				&Ours&0.025 $\pm$0.001&0.054 $\pm$0.001&{\bf 0.044 $\pm$0.002}\\
				\midrule
				\multirow{2}{*}{1W-A}&{\bf CILRS}&{\bf 0.025 $\pm$0.005}&{\bf 0.097 $\pm$0.006}&{\bf 0.116 $\pm$0.011}\\
				&Ours&0.041 $\pm$0.008&0.117 $\pm$0.016&0.126 $\pm$0.018\\
				\midrule
				\multirow{2}{*}{AW-N}&{\bf CILRS}&{\bf 0.024 $\pm$0.001}&{\bf 0.054 $\pm$0.008}&0.057 $\pm$0.011\\
				&Ours&0.029 $\pm$0.002&0.069 $\pm$0.005&{\bf 0.044 $\pm$0.005}\\
				\midrule
				\multirow{2}{*}{AW-A}&{\bf CILRS}&{\bf 0.024 $\pm$0.003}&{\bf 0.088 $\pm$0.006}&{\bf 0.102 $\pm$0.007}\\
				&Ours&0.035 $\pm$0.006&0.120 $\pm$0.009&0.107 $\pm$0.008\\
				\bottomrule
			\end{tabular}
		}
	\end{center}
	\label{tab:control_compare}
	\begin{tablenotes}\small
		\item $MAE_{ST}$: mean absolute error of steering prediction, $MAE_{TH}$: mean absolute error of throttle prediction, $MAE_{BR}$: mean absolute error of brake prediction.
		\item CILRS: Conditional Imitation Learning-based model\cite{cilrs} (R: using ResNet\cite{resnet}, S: with Speed input).
	\end{tablenotes}
\end{table}

\subsubsection{Vehicular Controls Estimation} \label{subss:control_compare}
Based on Table \ref{tab:control_compare}, especially in a comparison between the result in 1W-N with 1W-A and AW-N with AW-A, both CILRS and our model have larger MAE scores in the adversarial conditions. This means that both models have inferior performance due to various unexpected situations such as predicting the vehicular controls (steering, throttle, brake) when a pedestrian is crossing the street suddenly. This phenomenon also appears in the waypoints prediction performance evaluation meaning that adversarial driving is a big challenge for all models, especially on their controller module. Furthermore, our model loses to CILRS where it constantly has a larger MAE, especially on steering and throttle estimation. This result is as expected since CILRS is fed with one-hot encoded high-level navigational commands that can boost its confidence in estimating vehicular controls level. However, this result is contradictive to the result shown in Table \ref{tab:8T} where CILRS cannot perform very well in the driving task on CARLA simulator. This is because the model is only performing inference on a set of recorded driving data and has nothing to do with the environment state. Therefore, each prediction made by the model will not affect the future state of the environment which is critically affecting model performance. Moreover, as explained in Subsection \ref{subs:related2}, providing high-level commands is less reliable in real-world autonomous driving as there is no sensor that can give high-level commands other than the command from the driver itself. Therefore, although results in a larger MAE, providing GPS locations along with global to local coordinate transformation is preferable since they still can give an explicit intuition of navigational commands to the model which is useful for vehicular controls estimation.

\section{Conclusion} \label{sec:conclusion}
In this research, we present an end-to-end deep multi-task learning model to handle both perception and control tasks simultaneously for an autonomous driving vehicle. We consider point-to-point navigation task where the model is obligated to drive the ego vehicle by following a sequence of routes defined by the global planner. CARLA simulator with four different scenarios is used to evaluate the model and understand several aspects of driving. A comparative study with some recent models is conducted to justify model performance. Then, an ablation study is also conducted to understand the behavior and roles of some model functions. Furthermore, an extensive evaluation with task-wise performance scoring is conducted to analyze the model's capability in handling multiple tasks simultaneously.

Based on the experiment result, we disclosed several findings as follows. First, we conclude that all models suffer in adversarial situations and various weather conditions (except the AIM model in the AW-A scenario). Second, our model achieves the highest driving score (DS) as it can react properly to abnormalities thanks to the SDC mapping that provides stable features to the perception module. Third, our model can maintain the trade-off between route completion (RC) and infraction penalty (IP) as it understands different aspects of driving supported by two agents. Moreover, our model is more efficient as it has fewer trainable parameters and uses less GPU utilization compared to the runner-up of each scenario. Fourth, the AIM model has the highest RC, however, it cannot achieve the highest DS as it makes many infractions due to its less awareness. Fifth, the fusion-based models are too careful to drive and result in low RC. The LiDAR features are biasing too much so that the model loses important information. Sixth, based on the ablation study, SDC mapping and multi-agent are proven to play important roles in enhancing model drivability as they provide better perception and more control options. Lastly, based on the task-wise performance comparison, our model has better performances in waypoint prediction, traffic light state prediction, and stop sign prediction tasks. Although it loses in semantic segmentation and vehicular controls estimation tasks, our model is still preferable considering its size and reliability.

As for future studies, the research can be extended to the use of RGBD cameras in multiple views to capture more information. With a better perception, the model is expected to have better drivability as the controller is provided with more useful features. Furthermore, the model can be implemented on a real car or vehicular robot. However, noisy data and measurement errors need to be considered carefully as they can affect model performance and other functions.

\ifCLASSOPTIONcaptionsoff
  \newpage
\fi



\bibliographystyle{IEEEtran}
\bibliography{references}
%




%


\begin{IEEEbiography}[{\includegraphics[width=1in,height=1.25in,clip,keepaspectratio]{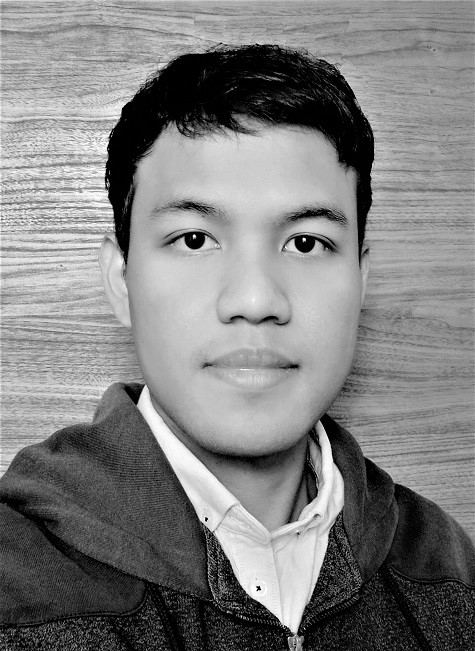}}]{Oskar Natan}
	received the B.A.Sc. degree in electronics engineering and the M.Eng. degree in electrical engineering from the Electronic Engineering Polytechnic Institute of Surabaya, Indonesia, in 2017 and 2019, respectively. He is currently pursuing the Dr.Eng. degree with the Department of Computer Science and Engineering, Toyohashi University of Technology, Japan. Since January 2020, he has been a Lecturer with the Department of Computer Science and Electronics, Gadjah Mada University, Indonesia. 
\end{IEEEbiography}




\begin{IEEEbiography}[{\includegraphics[width=1in,height=1.25in,clip,keepaspectratio]{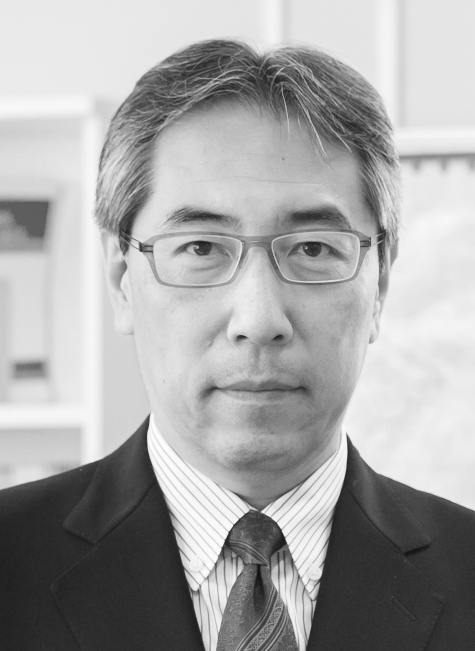}}]{Jun Miura}
	(Member, IEEE) received the B.Eng. degree in mechanical engineering and the M.Eng. and Dr.Eng. degrees in information engineering from The University of Tokyo, Japan, in 1984, 1986, and 1989, respectively. From 1989 to 2007, he was with the Department of Computer-Controlled Mechanical Systems, Osaka University, Japan, first as a Research Associate and later as an Associate Professor. From March 1994 to February 1995, he was a Visiting Scientist with the Department of Computer Science, Carnegie Mellon University, USA. In 2007, he became a Professor with the Department of Computer Science and Engineering, Toyohashi University of Technology, Japan. He has published over 245 scientific articles in the field of robotics and artificial intelligence in internationally reputable journals and conferences. He has received plenty of awards, including the Best Paper Award from the Robotics Society of Japan in 1997, the Best Paper Award Finalist at ICRA 1995, and the Best Service Robotics Paper Award Finalist at ICRA 2013. 
\end{IEEEbiography}



\vfill


\end{document}